\documentclass[10pt,twocolumn,letterpaper]{article}

\usepackage{cvpr}
\usepackage{times}
\usepackage{epsfig}
\usepackage{graphicx}
\usepackage{amsmath}
\usepackage{amssymb}
\usepackage{booktabs}

\usepackage[breaklinks=true,bookmarks=false]{hyperref}

\cvprfinalcopy 

\def\cvprPaperID{273} 

\begin{document}

\title{Conditional Similarity Networks}

\author{Andreas Veit$^1$
	\quad Serge Belongie$^1$ \quad Theofanis Karaletsos$^{2,3}$\\
	{ \{{av443, sjb344}\}{@cornell.edu, theofanis.karaletsos@gmail.com }} \\
	$^1$ Department of Computer Science \& Cornell Tech, Cornell University\\
		$^2$ Uber AI Labs, \ \ 
        $^3$ Computational Biology, Sloan Kettering Institute
	}

\maketitle

\begin{abstract}
What makes images similar? To measure the similarity between images, they are typically embedded in a feature-vector space, in which their distance preserve the relative dissimilarity. However, when learning such similarity embeddings the simplifying assumption is commonly made that images are only compared to one unique measure of similarity. A main reason for this is that contradicting notions of similarities cannot be captured in a single space. To address this shortcoming, we propose Conditional Similarity Networks (CSNs) that learn embeddings differentiated into semantically distinct subspaces that capture the different notions of similarities. CSNs jointly learn a disentangled embedding where features for different similarities are encoded in separate dimensions as well as masks that select and reweight relevant dimensions to induce a subspace that encodes a specific similarity notion. We show that our approach learns interpretable image representations with visually relevant semantic subspaces. Further, when evaluating on triplet questions from multiple similarity notions our model even outperforms the accuracy obtained by training individual specialized networks for each notion separately. 
\end{abstract}
\vspace{-10pt}
\section{Introduction}
Understanding visual similarities between images is a key problem in computer vision. To measure the similarity between images, they are embedded in a feature-vector space, in which their distances preserve the relative dissimilarity. Commonly, convolutional neural networks are trained to transform images into respective feature-vectors. We refer to these as Similarity Networks. When learning such networks from pairwise or triplet (dis-)similarity constraints, the simplifying assumption is commonly made that objects are compared according to one unique measure of similarity. However, objects have various attributes and can be compared according to a multitude of semantic aspects.

\begin{figure}[t]
\begin{center}
\includegraphics[width=0.7\linewidth]{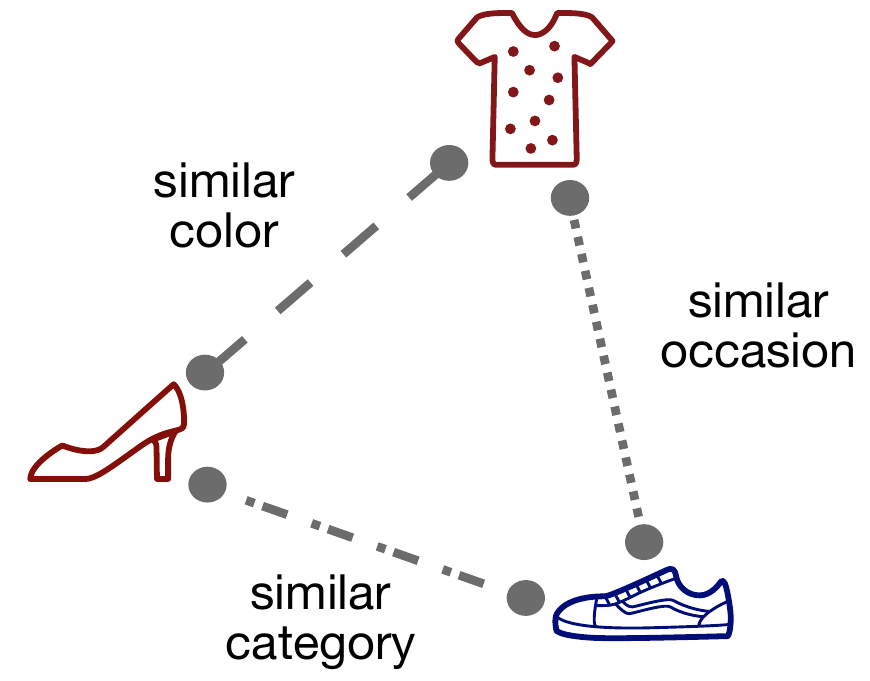}
\end{center}
\vspace{-5pt}
\caption{Example illustrating how objects can be compared according to multiple notions of similarity. Here, we demonstrate three intuitive concepts, which are challenging to combine for a machine vision algorithm that has to embed objects in a feature space where distances preserve the relative dissimilarity: shoes are of the same category; red objects are more similar in terms of color; sneakers and t-shirts are stylistically closer.}
\label{fig:fig1}
\vspace{-10pt}
\end{figure}

\begin{figure*}[t]
\begin{center}
\includegraphics[width=0.95\linewidth]{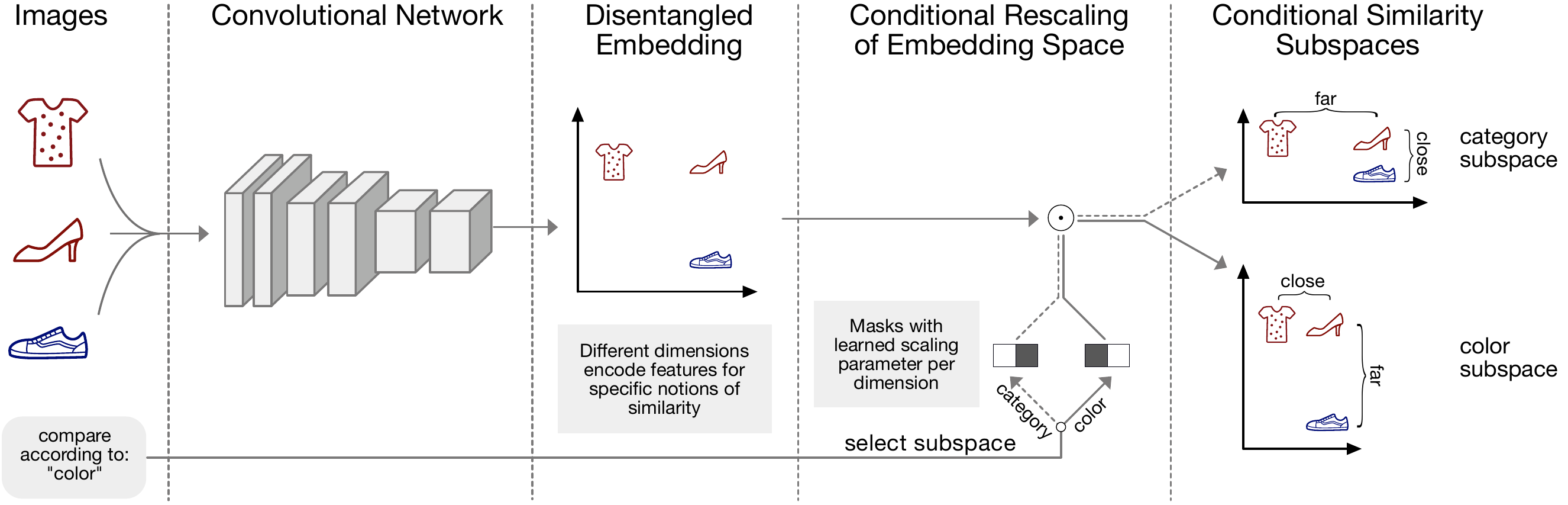}
\end{center}
\vspace{-5pt}
\caption{The proposed Conditional Similarity Network consists of three key components: First, a learned convolutional neural network as feature extractor that learns the disentangled embedding, i.e., different dimensions encode features for specific notions of similarity. Second, a condition that encodes according to which visual concept images should be compared. Third, a learned masking operation that, given the condition, selects the relevant embedding dimensions that induce a subspace which encodes the queried visual concept.}
\label{fig:overview}
\vspace{-10pt}
\end{figure*}

An illustrative example to consider is the comparison of coloured geometric shapes, a task toddlers are regularly exposed to with benefits to concept learning. Consider, that a red triangle and a red circle are very similar in terms of color, more so than a red triangle and a blue triangle. However, the triangles are more similar to one another in terms of shape than the triangle and the circle. 

An optimal embedding should minimize distances between perceptually similar objects. 
In the example above and also in the practical example in Figure~\ref{fig:fig1} this creates a situation where the same two objects are semantically repelled and drawn to each other at the same time. A standard triplet embedding ignores the sources of similarity and cannot jointly satisfy the competing semantic aspects. Thus, a successful embedding necessarily needs to take the visual concept into account that objects are compared to.

One way to address this issue is to learn separate triplet networks for each aspect of similarity. However, the idea is wasteful in terms of parameters needed, redundancy of parameters, as well as the associated need for training data. 

In this work, we introduce Conditional Similarity Networks (CSNs) a joint architecture to learn a nonlinear embeddings that gracefully deals with multiple notions of similarity within a shared embedding using a shared feature extractor. 
Different aspects of similarity are incorporated by assigning responsibility weights to each embedding dimension with respect to each aspect of similarity. This can be achieved through a masking operation leading to separate semantic subspaces. Figure~\ref{fig:overview} provides an overview of the proposed framework. Images are passed through a convolutional network and projected into a nonlinear embedding such that different dimensions encode features for specific notions of similarity. Subsequent masks indicate which dimensions of the embedding are responsible for separate aspects of similarity. We can then compare objects according to various notions of similarity by selecting an appropriate masked subspace. In the proposed approach the convolutional network that learns the disentangled embedding as well as the masks that learn to select relevant dimensions are trained jointly.

In our experiments we evaluate the quality of the learned embeddings by their ability to embed unseen triplets. We demonstrate that CSNs clearly outperform single triplet networks, and even sets of specialist triplet networks where a lot more parameters are available and each network is trained towards one single similarity notion. Further we show CSNs make the representation interpretable by encoding different similarities in separate dimensions.

Our contributions are 
a) formulating Conditional Similarity Networks, an approach that allows to to learn nonlinear embeddings that incorporate multiple aspects of similarity within a shared embedding using a shared feature extractor,
b) demonstrating that the proposed approach outperforms standard triplet networks and even sets of specialist triplet networks in a variety of hard predictive visual tasks and 
c) demonstrating that our approach successfully disentangles the embedding features into meaningful dimensions so as to make the representation interpretable.

\section{Related Work}
Similarity based learning has emerged as a broad field of interest in modern computer vision and has been used in many contexts. 
Disconnected from the input image, triplet based similarity embeddings, can be learned using crowd-kernels~\cite{van2012stochastic}. Further, Tamuz et al.~\cite{tamuz2011adaptively} introduce a probabilistic treatment for triplets and learn an adaptive crowd kernel. Similar work has been generalized to multiple-views and clustering settings by Amid and Ukkonen~\cite{amid2015multiview} as well as Van der Maaten and Hinton~\cite{van2012visualizing}. A combination of triplet embeddings with input kernels was presented by Wilber et al.~\cite{wilber2015learning}, but this work did not include joint feature and embedding learning. 
An early approach to connect input features with embeddings has been to learn image similarity functions through ranking~\cite{chechik2010large}.

A foundational line of work combining similarities with neural network models to learn visual features from similarities revolves around Siamese networks~\cite{chopra2005learning,hadsell2006dimensionality}, which use pairwise distances to learn embeddings discriminatively. In contrast to pairwise comparisons, triplets have a key advantage due to their flexibility in capturing a variety of higher-order similarity constraints rather than the binary similar/dissimilar statement for pairs. 
Neural networks to learn visual features from triplet based similarities have been used by Wang et al.~\cite{wang2014learning} and Schroff et al.~\cite{schroff2015facenet} for face verification and fine-grained visual categorization. A key insight from these works is that semantics as captured by triplet embeddings are a natural way to represent complex class-structures when dealing with problems of high-dimensional categorization and greatly boost the ability of models to share information between classes.

Disentangling representations is a major topic in the recent machine learning literature and has for example been tackled using Boltzmann Machines by Reed et al.~\cite{reed2014learning}. Chen et al.~\cite{chen2016infogan} propose information theoretical factorizations to improve unsupervised adversarial networks.
Within this stream of research, the work closest to ours is that of Karaletsos et al.~\cite{Karaletsos2016} on representation learning which introduces a joint generative model over inputs and triplets to learn a factorized latent space. However, the focus of that work is the generative aspect of disentangling representations and proof of concept applications to low-dimensional data. Our work introduces a convolutional embedding architecture that forgoes the generative pathway in favor of exploring applications to embed high-dimensional image data. We thus demonstrate that the generative interpretation is not required to reap the benefits of Conditional Similarity Networks and demonstrate in particular their use in common computer vision tasks.

A theme in our work is the goal of modeling separate similarity measures within the same system by factorizing (or {\it disentangling}) latent spaces. We note the relation of these goals to a variety of approaches used in representation learning.  Multi-view learning~\cite{su2015multi,wang2015deep} has been used for 3d shape inference and shown to generically be a good way to learn factorized latent spaces. Multiple kernel learning~\cite{bach2004multiple,sonnenburg2006large} employs information encoded in different kernels to provide predictions using the synthesized complex feature space and has also been used for similarity-based learning by McFee and Lanckriet~\cite{mcfee2011learning}. Multi-task learning approaches~\cite{collobert2008unified} are used when information from disparate sources or using differing assumptions can be combined beneficially for a final prediction task. Indeed, our gating mechanism can be interpreted as an architectural novelty in neural networks for \emph{multi-task triplet learning}.
Similar to our work, multiliniear networks ~\cite{lin2015bilinear} also strive to factorize representations, but differ in that they ignore weak additional information.
An interesting link also exists to multiple similarity learning~\cite{babenko2009similarity}, where category specific similarities are used to approximate a fine-grained global embedding. Our global factorized embeddings can be thought of as an approach to capture similar information in a shared space directly through feature learning.

We also discuss the notion of attention in our work, by employing gates to attend to the mentioned subspaces of the inferred embeddings when focusing on particular visual tasks. This term may be confused with spatial attention such as used in the DRAW model~\cite{gregor2015draw}, but bears similarity insofar as it shows that the ability to gate the focus of the model on relevant dimensions (in our case in latent space rather than observed space) is beneficial both to the semantics and to the quantitative performance of our model.

\section{Conditional Similarity Networks}
Our goal is to learn a nonlinear feature embedding $f(x)$, from an image $x$ into a feature space $\mathbb{R}^d$, such that for a pair of images $x_1$ and $x_2$, the Euclidean distance between $f(x_1)$ and $f(x_2)$ reflects their semantic dis-similarity. In particular, we strive for the distance between images of semantically similar objects to be small, and the distance between images of semantically different objects to be large. This relationship should hold independent of imaging conditions.

We consider $y= f(x)$ to be an embedding of observed images $x$ into coordinates in a feature space $y$. Here, $f(x)= W g(x)$ clarifies that the embedding function is a composition of an arbitrarily nonlinear function $g(\cdot)$ and a linear projection $W$, for $W \in \mathbb{R}^{d \times b}$, where $d$ denotes the dimensions of the embedding and $b$ stands for the dimensions of the output of the nonlinear function $g(\cdot)$.
In general, we denote the parameters of function $f(x)$ by $\theta$, denoting all the filters and weights.

\subsection{Conditional Similarity Triplets}
\label{qdT}
Apart from observing images {\bf x}, we are also given a set of triplet constraints sampled from an oracle such as a crowd. We define triplet constraints in the following.
\begin{figure}[t]
\begin{center}
\includegraphics[width=1\linewidth]{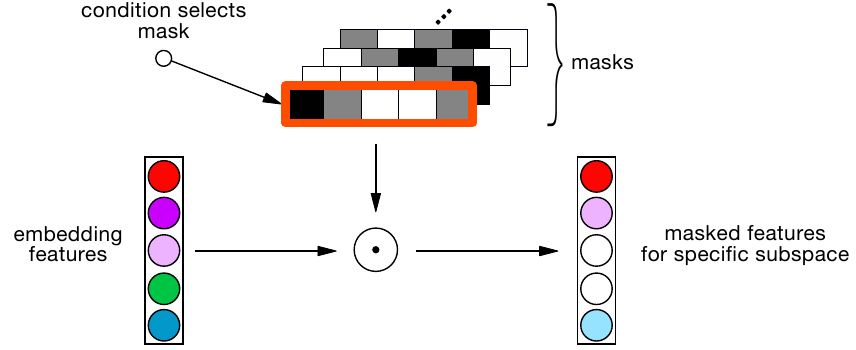}
\end{center}
\vspace{-5pt}
\caption{The masking operation selects relevant embedding dimensions, given a condition index. Masking can be seen as a soft gating function, to attend to a particular concept.}
\label{fig:mask_detail}
\vspace{-10pt}
\end{figure}

Given an unknown conditional similarity function $s_{c}(\cdot,\cdot)$, an oracle such as a crowd can compare images $x_1$, $x_2$ and $x_3$ according to condition $c$. A condition is defined as a certain notion of similarity according to which images can be compared. Figure~\ref{fig:fig1} gives a few example notions according to which images of fashion products can be compared. The condition $c$ serves as a switch between attented visual concepts and can effectively gate between different similarity functions $s_{c}$. Using image $x_1$ as reference, the oracle can apply $s_{c}(x_1,x_2)$ and $s_{c}(x_1,x_3)$ and decide whether $x_1$ is more similar to $x_2$ or to $x_3$ conditioned on $c$. The oracle then returns an ordering over these two distances, which we call a triplet $t$. A triplet is defined as the set of indices $\{\text{reference image, more distant image, closer image} \}$, e.g.\ $\{1, 3, 2\}$ if $s_{c}(x_1,x_3)$ is larger than $s_{c}(x_1,x_2)$.

We define the set of all triplets related to condition $C$ as:
\begin{align}
\label{triplet_definition}
\mathcal{T_{c}}=\{(i,j,l;c)~~|~~s_{c}(x_i,x_j) > s_{c}(x_i,x_l) \}.
\end{align}
We do not have access to the exhaustive set $\mathcal{T_{c}}$, but can sample $K$-times from it using the oracle to yield a finite sample $\mathcal{T_{c}}^{K}=\{t_k\}_{k=1}^{K}$.

\subsection{Learning From Triplets}
\label{triplet_loss}
The feature space spanned by our model is given by function $f(\cdot)$. To learn this nonlinear embedding and to be consistent with the observed triplets, we define a loss function $L_{T}(\cdot)$ over triplets to model the similarity structure over images. 
The triplet loss commonly used is
\begin{equation}
\label{eqn:standardtripletlearning}
\begin{split}
L_{T}(x_i,x_j,x_l)= \text{max}\{ 0, D(x_i,x_j)- D(x_i,x_l) + h  \}
\end{split}
\end{equation}
where $D(x_i,x_j) = \|f(x_i ; \theta)-f(x_j ; \theta)\|_2.$ is the Euclidean distance between the representations of images $x_i$ and $x_j$. The scalar margin $h$ helps to prevent trivial solutions. The generic triplet loss is not capable of capturing the structure induced by multiple notions of similarities.

To be able to model conditional similarities, we introduce masks ${\bf m}$ over the embedding with $m \in \mathbb{R}^{d \times n_c }$ where $n_c$ is the number of possible notions of similarities. We define a set of parameters $\beta_m$ of the same dimension as ${\bf m}$ such that ${\bf m}= \sigma(\beta)$, with $\sigma$ denoting a rectified linear unit so that $\sigma(\beta) = \max\{0,\beta\}$. As such, we denote $m_c$ to be the selection of the $c$-th mask column of dimension $d$ (in pseudocode $m_c={\bf m}[:,c]$). The mask plays the role of an element-wise gating function selecting the relevant dimensions of the embedding required to attend to a particular concept.  The role of the masking operation is visually sketched in Figure~\ref{fig:mask_detail}. The masked distance function between two images $x_i$ and $x_j$ is given by
\begin{align}
\label{query_aware_distance}
D(x_i,x_j;m_c,\theta) = \|f(x_i ; \theta) m_c-f(x_j ; \theta) m_c\|_2.
\end{align}
While appearing to be a small technical change, the inclusion of a masking mechanism for the triplet-loss has a highly non-trivial effect.
The mask induces a subspace over the relevant embedding dimensions, effectively attending only to the relevant dimensions for the visual concept being queried. 
In the loss function above, that translates into a modulated cost phasing out Euclidean distances between irrelevant feature-dimensions while preserving the loss-structure of the relevant ones. 

Given an triplet $t=\{i,j,l\}$ defined over indices of the observed images and a corresponding condition-index $c$, the final triplet loss function $\mathcal{L}_{T}(\cdot)$ is given by:
\begin{equation}
\label{eqn:triplet_loss}
\begin{split}
&\mathcal{L}_{T}(x_i,x_j,x_l, c ; m,\theta)= \\
&\text{max}\{ 0, D(x_i,x_j;m_c,\theta)- D(x_i,x_l;m_c,\theta) + h  \}
\end{split}
\end{equation}

\subsection{Encouraging Regular Embeddings}
We want to encourage embeddings to be drawn from a unit ball to maintain regularity in the latent space. We encode this in an embedding loss function $\mathcal{L}_{W}$ given by:
\begin{equation}
\mathcal{L}_{W}({\bf x} ; \theta) = {\|{f({\bf x} ; \theta)}\|}_2^2 = {\|{\bf y}\|}_2^2
\end{equation}
The separate subspaces are computed as $f(x)m_{c}$. To prevent the masks from expanding the embedding and to encourage sparse masks, we add a loss to regulate the masks:
\begin{equation}
\mathcal{L}_{M}({\bf m}) = {\|{\bf m}\|}_1
\end{equation}
Without these terms, an optimization scheme may choose to inflate embeddings to create space for new data points instead of learning appropriate parameters to encode the semantic structure.

\begin{figure}[t]
\begin{center}
\includegraphics[width=1\linewidth]{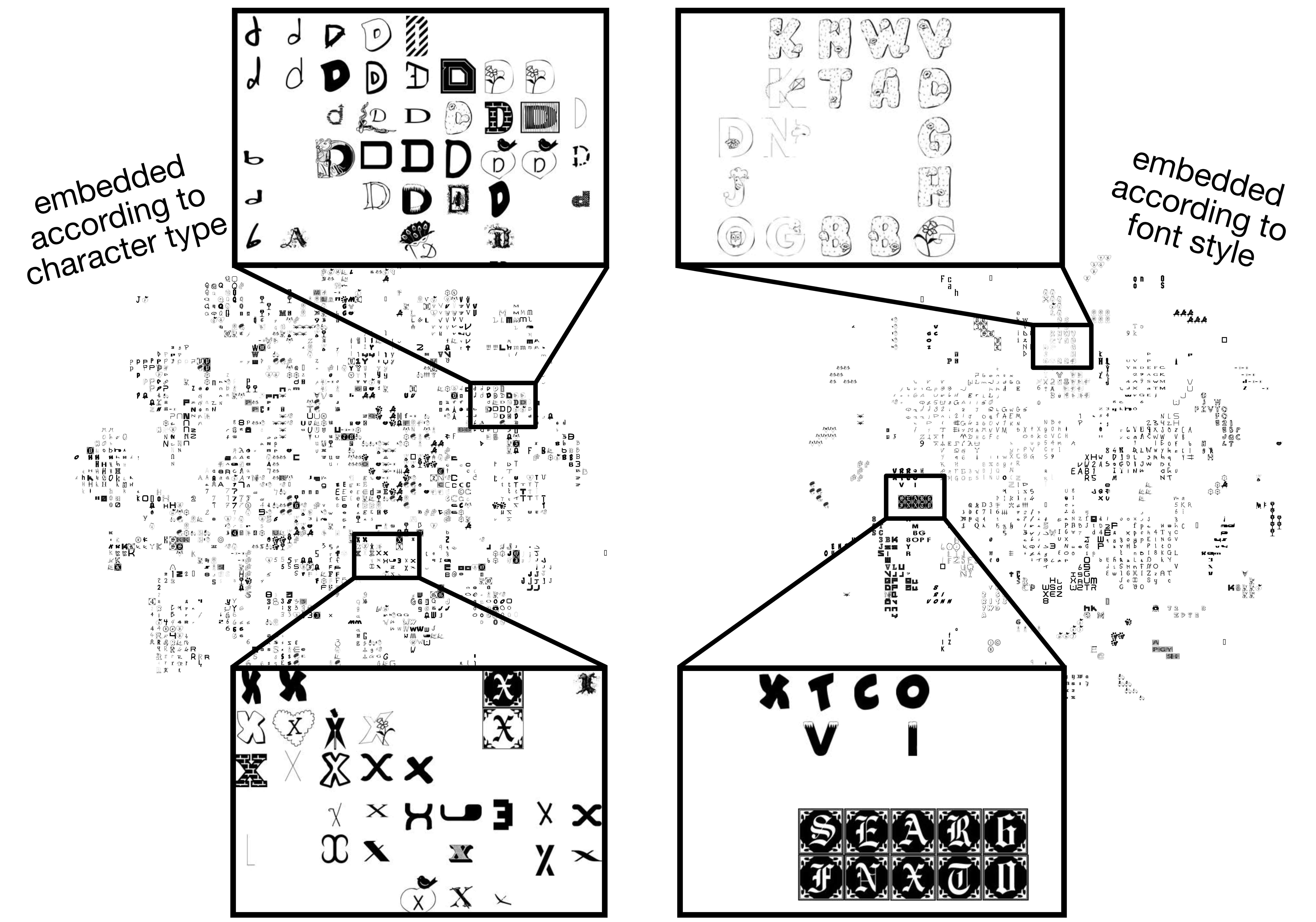}
\end{center}
\vspace{-5pt}
\caption{Visualization of 2D embeddings of two learned subspaces of the character feature space. The subspaces are obtained by attending to different subsets of the dimensions in the image representations. The subspace on the left groups images by character type, the one on the right according to font style. For clear visual representation we discretize the space into a grid and pick one image from each cell at random.}
\label{fig:fontembeddings}
\vspace{-10pt}
\end{figure}
\subsection{Joint Formulation For Convolutional CSNs}
We define a loss-function $\mathcal{L}_{CSN}$ for training CSNs by putting together the defined loss functions. Given images ${\bf x}$, triplet constraints with associated condition ${\bf \{t,\ c\}}$ as well as parameters for the masks ${\bf m}$ and the embedding function $\theta$, the CSN loss is defined as

\begin{equation}
\begin{split}
&\mathcal{L}_{CSN}({\bf x}, {\bf \{t,c\}};{\bf m},\theta) =\\ 
&\mathcal{L}_{T}(x_{t_0},x_{t_1},x_{t_2}, c ;{\bf m}, \theta) + \lambda_1 \mathcal{L}_{W}({\bf x},\theta)+ \lambda_2 \mathcal{L}_{M}({\bf m})
\end{split}
\end{equation}
The parameters $\lambda_1$ and $\lambda_2$ weight the contributions of the triplet terms against the regular embedding terms.

In our paper, the nonlinear embedding function is defined as $f(x)= W g(x)$, where $g(x)$ is a convolutional neural network. In the masked learning procedure the masks learn to select specific dimensions in the embedding that are associated with a given notion of similarity. At the same time, $f(\cdot)$ learns to encode the visual features such that different dimensions in the embedding encode features associated to specific semantic notions of similarity. 
Then, during test time each image can be mapped into this embedding by $f(\cdot)$. By looking at the different dimensions of the image's representation, one can reason about the different semantic notions of similarity. We call a feature space spanned by a function with this property ${\it disentangled}$, as it preserves the separation of the similarity notions through test time.

\section{Experiments}
We focus our experiments on evaluating the semantic structure of the learned embeddings and their subspaces as well as the underlying convolutional filters.

\begin{figure}[t]
\begin{center}
\begin{tabular}{@{}c@{}}
\includegraphics[width=1\linewidth]{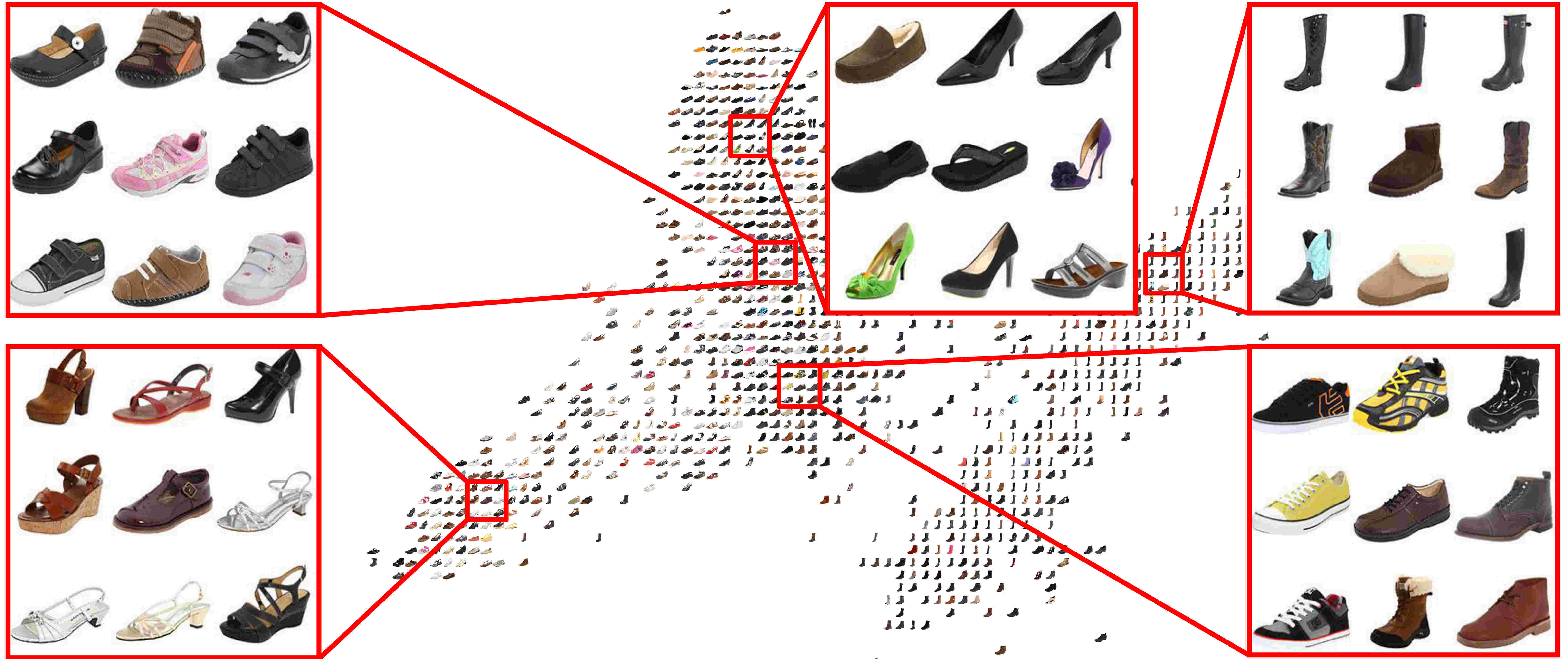} \\
(a) Embedding according to the closure mechanism \\
\includegraphics[width=1\linewidth]{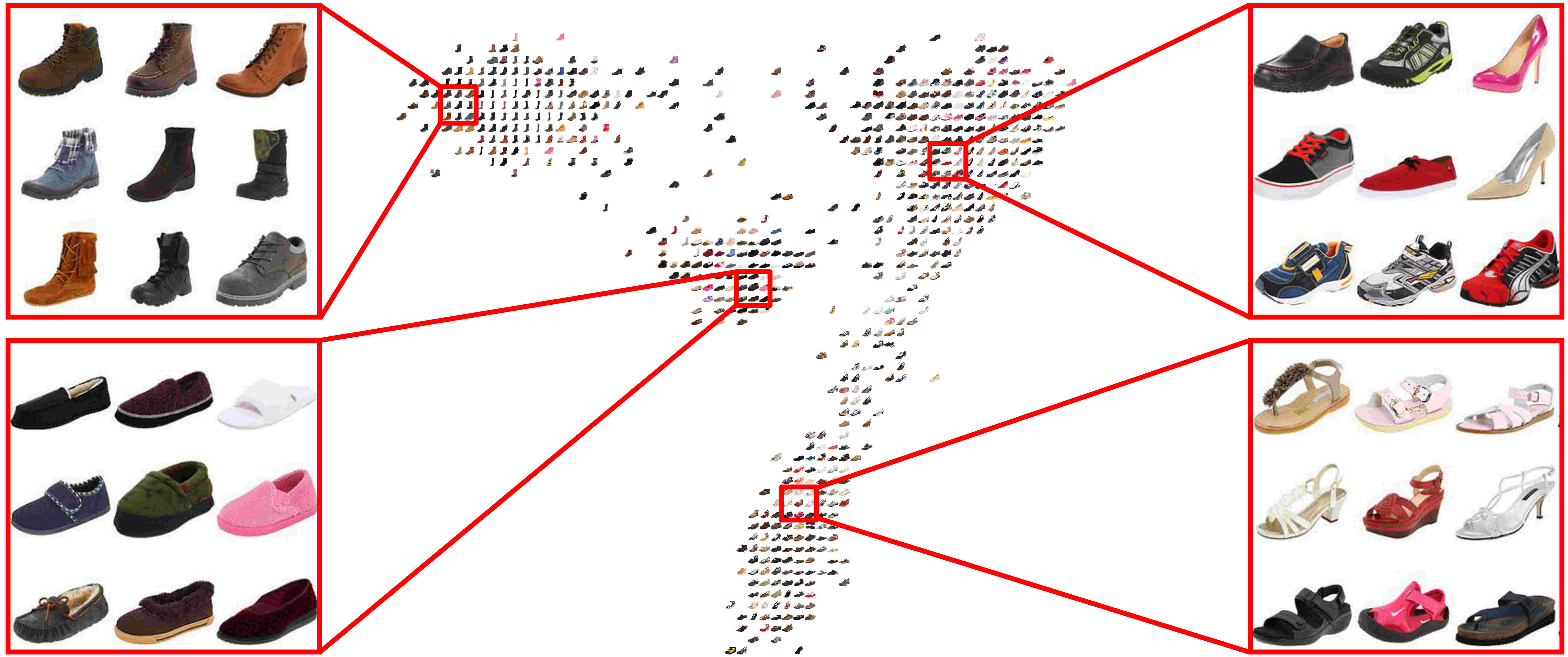} \\
(b) Embedding groups of boots, slippers, shoes and sandals\\
\end{tabular}
\end{center}
\vspace{-5pt}
\caption{Visualization of 2D embeddings of subspaces learned by the CSN. The spaces are clearly organized according to \textbf{(a)} closure mechanism of the shoes and \textbf{(b)} the category of the shoes. This shows that CSNs can successfully separate the subspaces.}
\label{fig:shoeembeddings1}
\vspace{-10pt}
\end{figure}
\subsection{Datasets}
We perform experiments on two different datasets. First, for illustrative purposes we use a dataset of fonts\footnote{http://erikbern.com/2016/01/21/analyzing-50k-fonts-using-deep-neural-networks/} collected by Bernhardsson. The dataset contains 3.1 million images of single characters in gray scale with a size of 64 by 64 pixels each. The dataset exhibits variations according to {\it font style} and {\it character type}. In particular, it contains 62 different characters in 50,000 fonts, from which we use the first 1,000.
Second, we use the Zappos50k shoe dataset~\cite{zappos50k} collected by Yu and Grauman. The dataset contains 50,000 images of individual richly annotated shoes, with a size of 136 by 102 pixels each, which we resize to 112 by 112. The images exhibit multiple complex variations. In particular, we are looking into four different characteristics: the {\it type of the shoes} (i.e., shoes, boots, sandals or slippers), the {\it suggested gender of the shoes} (i.e., for women, men, girls or boys), the {\it height of the shoes' heels} (numerical measurements from 0 to 5 inches) and the {\it closing mechanism of the shoes} (buckle, pull on, slip on, hook and loop or laced up). We also use the shoes' brand information to perform a fine-grained classification test.

To supervise and evaluate the triplet networks, we sample triplet constraints from the annotations of the datasets. For the font dataset, we sample triplets such that two characters are of the same type or font and one is different. For the Zappos dataset, we sample triplets in an analogous way for the three categorical attributes. For the heel heights we have numerical measurements so that for each triplet we pick two shoes with similar height and one with different height. First, we split the images into three parts: 70\% for training, 10\% for validation and 20\% in the test set. Then, we sample triplets within each set. For each attribute we collect 200k train, 20k validation and 40k test triplets.

\subsection{Baselines and Model Variants}
\begin{figure}[t]
\begin{center}
\includegraphics[width=1\linewidth]{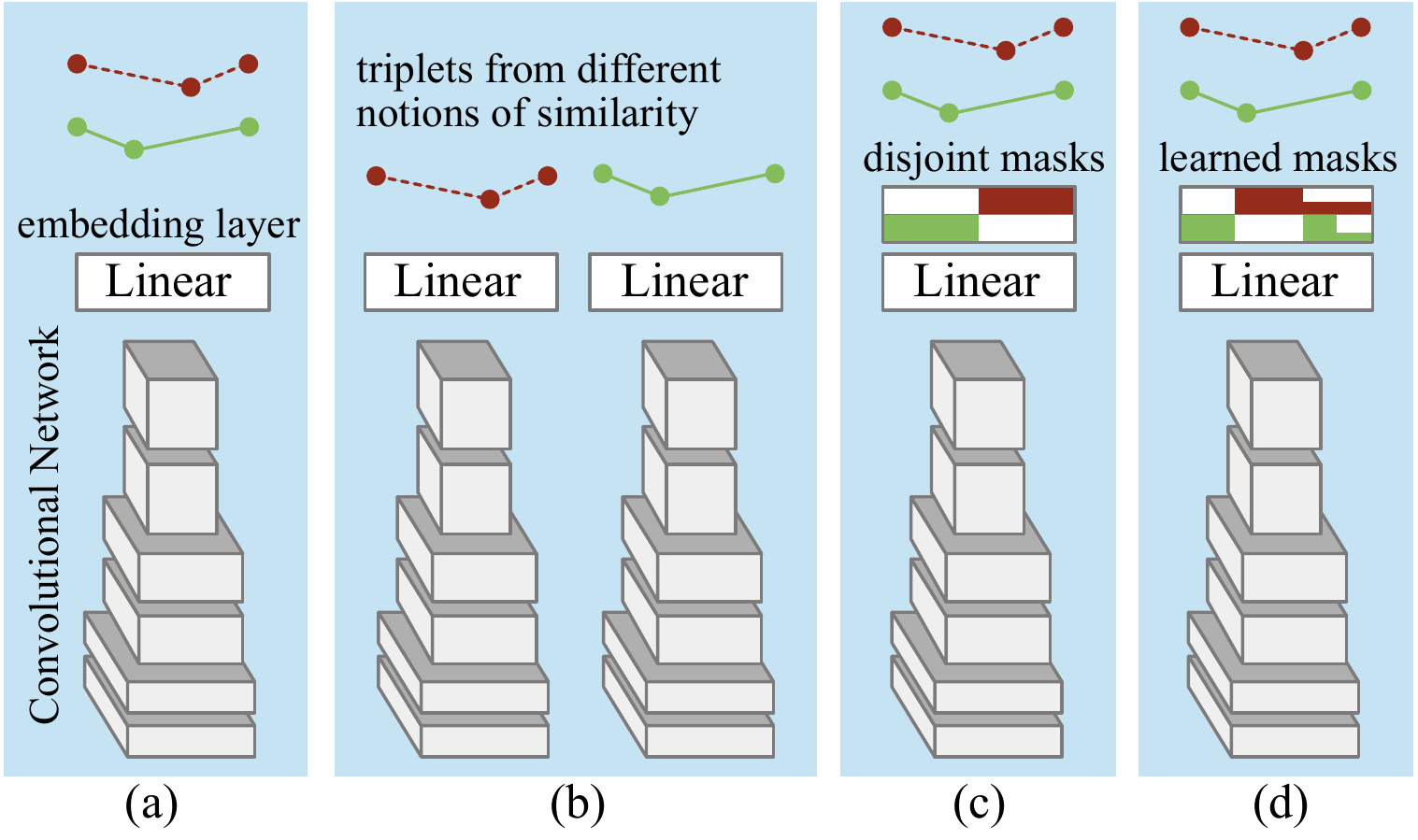}
\end{center}
\vspace{-10pt}
\caption{We show the four different model variants used in our experiments with the example of three objects being compared according to two contradictory notions of similarity, green and red. 
({\bf a}) A standard triplet network that treats all triplets equally ({\bf b}) A set of $n_c$-many triplet network experts specialized on green or red, respectively ({\bf c}) 
A CSN with masks pre-set to be disjoint, so that in the embedding each dimension encodes a feature for a specific notion of similarity ({\bf d}) A learned CSN, where the masks are learned to select features relevant to the respective notion of similarity.}
\label{fig:models}
\vspace{-10pt}
\end{figure}

As initial model for our experiments we use a ConvNet pre-trained on ImageNet. All model variants are fine-tuned on the same set of triplets and only differ in the way they are trained. We compare four different approaches, which are schematically illustrated in Figure~\ref{fig:models}.

\noindent
\textbf{Standard Triplet Network}: The common approach to learn from triplet constrains is a single Convolutional Network where the embedding layer receives supervision from the triplet loss defined in Equation~\ref{eqn:standardtripletlearning}. As such, it aims to learn from all available triplets jointly as if they come from a single measure of similarity. 

\noindent
\textbf{Set of Task Specific Triplet Networks}: Second, we compare to a set of $n_{c}$ separate triplet network experts, each of which is trained on a single notion of similarity. This overcomes the simplifying assumption that all comparisons come from a single measure of similarity. However, this comes at the cost of significantly more parameters. This is the best model achievable with currently available methods.

\noindent
\textbf{Conditional Similarity Networks - fixed disjoint masks}: We compare two variants of Conditional Similarity Networks. Both extend a standard triplet network with a masking operation on the embedding vector and supervise the network with the loss defined in Equation~\ref{eqn:triplet_loss}. 
The first variant learns the convolutional filters and the embedding. The masks are pre-defined to be disjoint between the different notions of similarity. 
This ensures the learned embedding is fully disentangled, because each dimension must encode features that describe a specific notion of similarity.

\noindent
\textbf{Conditional Similarity Networks - learned masks}: 
The second variant learns the convolutional filters, the embedding and the mask parameters together. This allows the model to learn unique features for the subspaces as well as features shared across tasks.
This variant has the additional benefit that the learned masks can provide interesting insight in how different similarity notions are related.

\subsection{Training Details}
\noindent
We train different convolutional networks for the two datasets. For the font dataset, we use a variant of the VGG architecture~\cite{simonyan2014very} with 9 layers of 3 by 3 convolutions and two fully connected layers, which we train from scratch. For the Zappos dataset we fine-tune an 18~layer deep residual network~\cite{he2015deep} that is pre-trained on Imagenet~\cite{deng2009imagenet}. We remove one downsampling module to adjust for the smaller image size. We train the networks with a mini-batch size of 256 and optimize using ADAM~\cite{kingma2014adam} with $\alpha = 5\text{\sc{e}-}5$, $\beta_1 = 0.1$ and $\beta_2 = 0.001$. For all our experiments we use an embedding dimension of 64 and the weights for the embedding losses are $\lambda_1 = 5\text{\sc{e}-}3$ and $\lambda_2 = 5\text{\sc{e}-}4$. In each mini-batch we sample triplets uniformly and for each condition in equal proportions. We train each model for 200 epochs and perform early stopping in that we evaluate the snapshot with highest validation performance on the test set.  

For our CSN variants, we use two masks over the embedding for the fonts dataset and four masks for the Zappos dataset, one mask per similarity notion. For models with pre-defined masks, we allocate $1/n_c$ th of the embedding dimensions to one task. When learning masks, we initialize $\beta_m$ using a normal distribution with 0.9 mean and 0.7 variance. 
Following the ReLU, this results in initial mask values that induce random subspaces for each similarity measure. We observe that different random subspaces perform better than a setup where all subspaces start from the same values. Masks that are initialized as disjoint analogous to the pre-defined masks perform similar to random masks, but are not able to learn shared features.

\subsection{Visual Exploration of the Learned Subspaces}
We visually explore the learned embeddings regarding their consistency according to respective similarity notions. We stress that all of these semantic representations are taking place within a shared space produced by the same network. The representations are {\it disentangled} so that each dimension encodes a feature for a specific notion of similarity. This allows us to use a simple masking operation to look into a specific semantic subspace.

\begin{figure}[t]
\begin{center}
\begin{tabular}{@{}c@{}}
\includegraphics[width=1\linewidth]{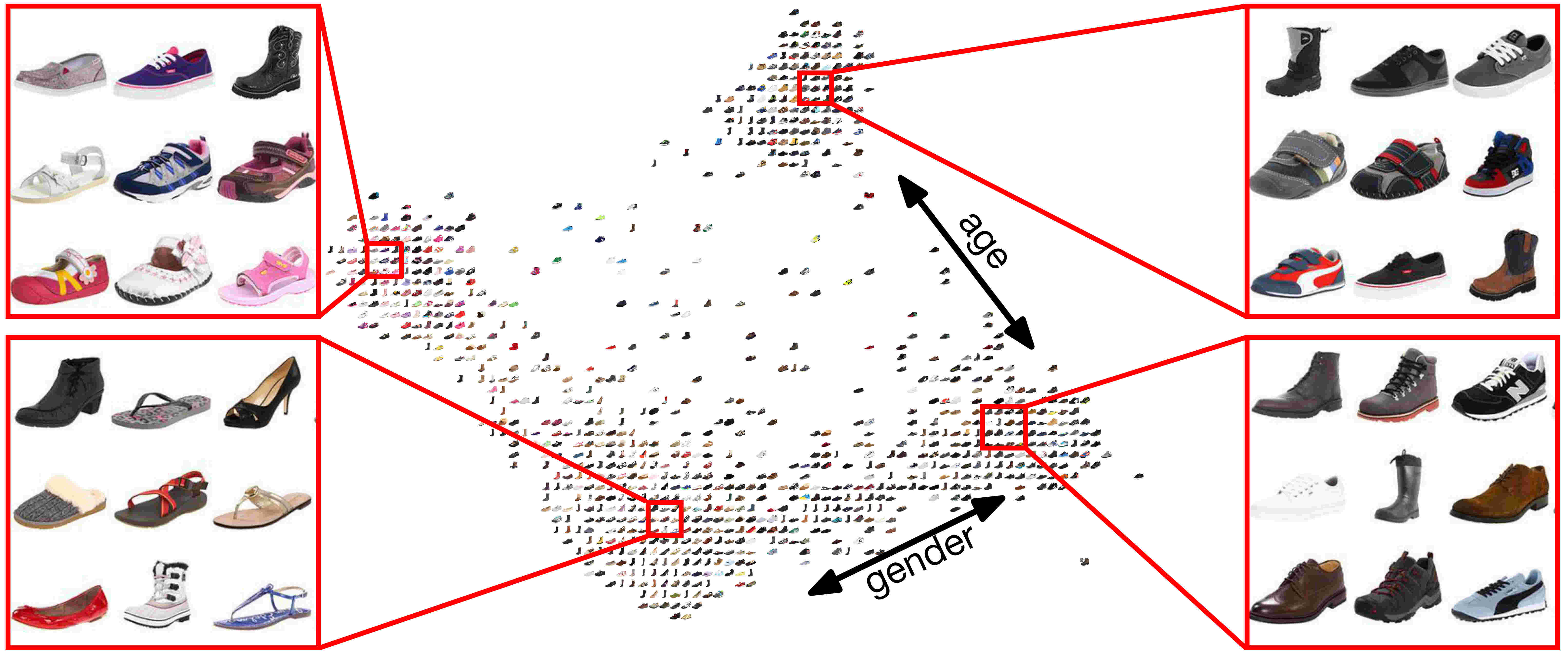} \\
(a) Embedding according to the suggested gender\\
\includegraphics[width=1\linewidth]{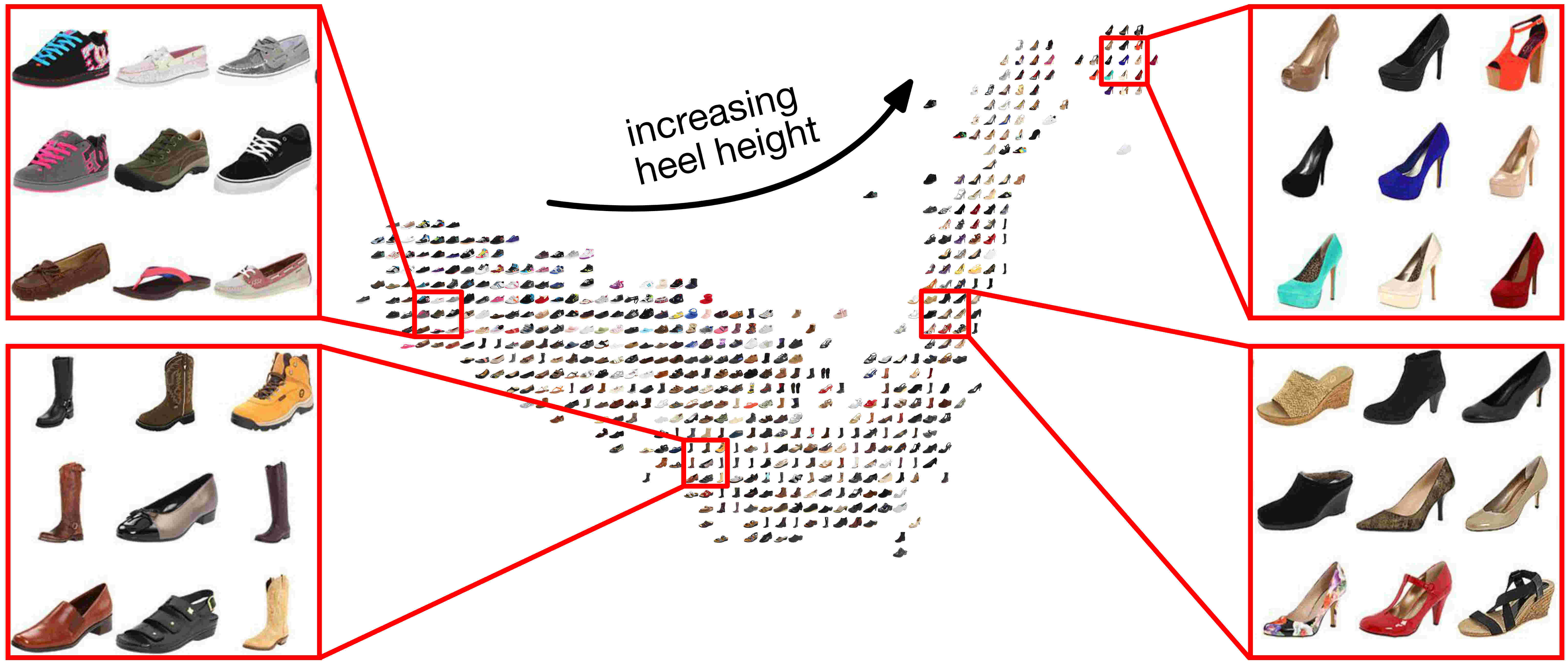} \\
(b) Embedding according to the height of the heels
\end{tabular}
\end{center}
\vspace{-5pt}
\caption{Visualization of the subspaces according to \textbf{(a)} suggested gender for the shoes and \textbf{(b)} height of the shoes' heel. The result shows that CSNs can learn categorical as well as continuous characteristics at the same time.}
\label{fig:shoeembeddings2}
\vspace{-10pt}
\end{figure}

Figure~\ref{fig:fontembeddings} shows embeddings of the two subspaces in the Fonts dataset, which we project down to two dimensions using t-SNE~\cite{van2008visualizing}. The learned features are successfully disentangled such that the dimensions selected by the first mask describe the character type (left) and those selected by the second mask the font style (right).
Figures~\ref{fig:shoeembeddings1} and~\ref{fig:shoeembeddings2} show embeddings of the four subspaces learned with a CSN on the Zappos50k dataset. Figure~\ref{fig:shoeembeddings1}(a) shows the subspace encoding features for the {\it closure mechanism} of the shoes. Figure~\ref{fig:shoeembeddings1}(b) shows the subspace attending to the {\it type} of the shoes. The embedding clearly separates the different types of shoes into boots, slippers and so on. Highlighted areas reveal some interesting details. For example, the highlighted region on the upper right side shows nearby images of the same type ('shoes') that are completely different according to all other aspects. This means the selected feature dimensions successfully focus only on the \emph{type} aspect and do not encode any of the other notions.
Figure~\ref{fig:shoeembeddings2}(a) shows the subspace for {\it suggested gender} for the shoes. The subspace separates shoes that are for female and male buyers as well as shoes for adult or youth buyers. The learned submanifold occupies a rotated square with axes defined by gender and age.
Finally, Figure~\ref{fig:shoeembeddings2}(b) shows a continuous embedding of {\it heel heights}, which is a subtle visual feature.

\subsection{Qualitative Analysis Of Subspaces}
\begin{figure}[t]
\begin{center}
\includegraphics[width=1\linewidth]{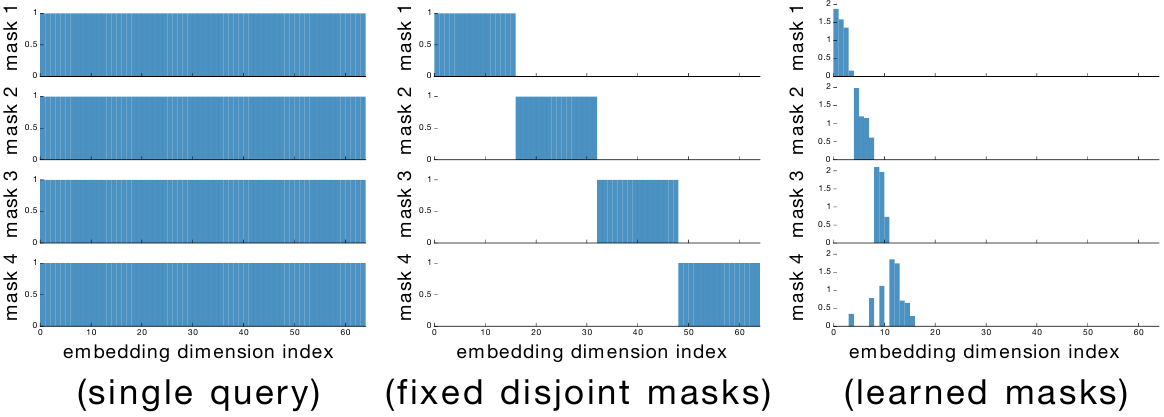}
\end{center}
\vspace{-5pt}
\caption{Visualization of the masks: \textbf{Left:} In standard triplet networks, each dimension is equally taken into account for each triplet. \textbf{Center:} The Conditional Similarity Network allows to focus on a subset of the embedding to answer a triplet question. Here, each mask focuses on one fourth. \textbf{Right:} For learned masks, it is evident that the model learns to switch off different dimensions per question. Further, a small subset is shared across tasks.}
\label{fig:masks}
\vspace{-10pt}
\end{figure}

The key feature of CSNs is the fact that they can learn separated semantic subspaces in the embeddings using the masking mechanism. We visualize the masks for our common model choices in Figure~\ref{fig:masks}. We show the traditional triplet loss, where each dimension is equally taken into account for each triplet. Further, we show pre-defined masks that are used to factorize the embedding into fully disjoint features. Lastly, we show a learned mask. Interestingly, the masks are very sparse in accordance with the 2D embeddings presented in the previous section, confirming that the concepts are low-dimensional. Further, although many additional dimensions are available, the model learned to share some of the features across concepts. This demonstrates that CSNs can learn to only use the required number of dimensions via relevance determination, reducing the need for picking the right embedding dimensionality.

\subsection{Results on Triplet Prediction}
\label{quant_triplet_pred}
To evaluate the quality of the learned embeddings by the different model variants, we test how well they generalize to unseen triplets. In particular, we perform triplet prediction on a testset of hold-out triplets from the Zappos50k dataset. We first train each model on a fixed set of triplets, where triplets are sourced from the four different notions of similarity. After convergence, we evaluate for each triplet with associated query $\{i,j,l,c\}$ in the testset whether the distance between $i$ and $l$ is smaller than between $i$ and $j$ according to concept/query $c$. Since this is a binary task, random guessing would perform at an error rate of 50\%.

The error rates for the different models are shown in Table~\ref{tab:results}. Standard Triplet Networks fail to capture fine-grained similarity and only reach an error rate of $23.72\%$. The set of task specific triplet networks greatly improves on that, achieving an error rate of $11.35\%$. This shows that simply \emph{learning a single space cannot capture multiple similarity notions}. However, this comes at a the cost of $n_c$ times more model parameters. 
Conditional Similarity Networks with fixed disjoint masks achieve an error rate of $10.79\%$, clearly outperforming both the single triplet network as well as the set of specialist networks, which have a lot more parameters available for learning. 
This means by factorizing the embedding space into separate semantic subspaces, CSNs can successfully capture multiple similarity notions without requiring substantially more parameters. Moreover, CSNs benefit from learning all concepts jointly within one model, utilizing shared structure between the concepts while keeping the subspaces separated. 
CSNs with learned masks achieve an error rate of $10.73\%$ improving performance even further. This indicates the benefits from allowing the model to determine the relevant dimensions and to share features across concepts.

\begin{table}[h]
\centering
\caption{\label{tab:results}Triplet Prediction Results: We evaluate how many triplets of the test set are satisfied in the learned embeddings. Triplets come from four different similarity notions. The proposed Conditional Similarity Network clearly outperforms standard triplet networks that treat each triplet as if it came from the same similarity notion. Moreover, CSNs even outperform sets of specialist triplet networks where a lot more parameters are available during training and each network is specifically trained towards one similarity notion. CSNs with learned masks provide the best performance.}
\begin{tabular}{@{}ll@{}} \toprule
    \text{Method} & \text{Error Rate} \\ \midrule
    \text{Standard Triplet Network}  & 23.72\% \\ 
    \text{Set of Specialized Triplet Networks}  & 11.35\%\\
    \text{CSN fixed disjoint masks}  & {10.79\%} \\ 
    \textbf{CSN learned masks}  & \textbf{10.73\%}\\\bottomrule
\end{tabular}
\end{table}

\begin{figure}[t]
\begin{center}
\includegraphics[width=0.8\linewidth]{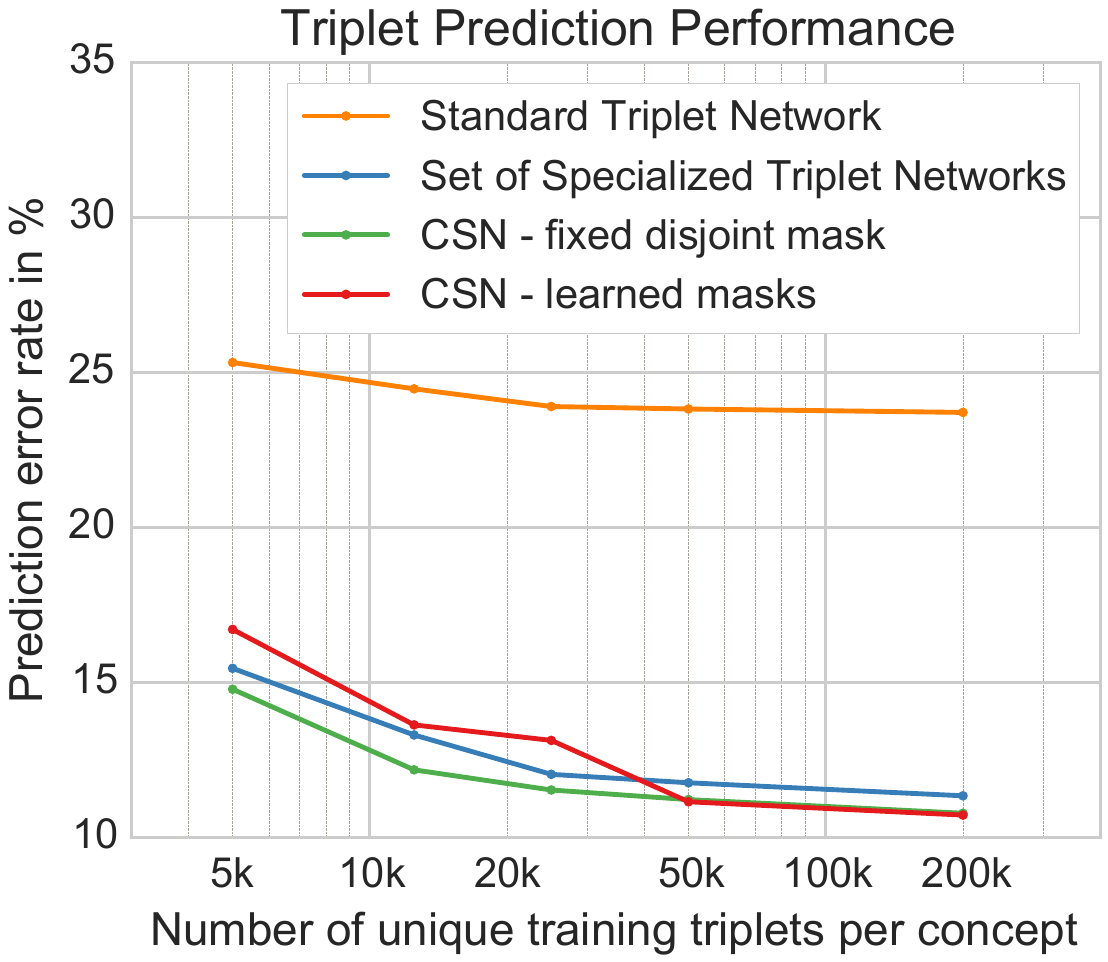}
\end{center}
\vspace{-5pt}
\caption{Triplet prediction performance with respect to number of unique training triplets available. CSNs with fixed masks consistently outperform the set of specialized triplet networks. CSNs with learned masks generally require more triplets, since they need to learn the embedding as well as the masks. However, when enough triplets are available, they provide the best performance.}
\label{fig:prediction}
\vspace{-10pt}
\end{figure}

Further, we evaluate the impact of the number of unique triplets available during training on performance. We compare models trained on $5$, $12.5$ $25$, $50$ and $200$ thousand triplets per concept. Figure~\ref{fig:prediction} shows that triplet networks generally improve with more available triplets. Further, CSNs with fixed masks consistently outperform set of specialized triplet networks. Lastly, CSNs with learned masks generally require more triplets, since they need to learn the embedding as well as the masks. However, when enough triplets are available, they provide the best performance.

\subsection{Analysis of Convolutional Features Using Off-Task Classification}
We now evaluate how the different learning approaches affect the visual features of the networks. We compare standard triplet networks to CSNs. Both are initialized from the same ImageNet pre-trained residual network and fine-tuned using the same triplets and with their respective losses as described in Section~\ref{quant_triplet_pred}. We evaluate the features learned by the two approaches, by subsequently performing brand classification on the Zappos dataset. In particular, we keep all convolutional filters fixed and replace the last embedding layer for both networks with one hidden and one softmax classification layer. We select the 30 brands in the Zappos dataset with the most examples and train with a standard multi-class classification approach using the 30 brands as classes. It is noteworthy that the triplets used for the fine-tuning do not contain brand information.
 
\begin{table}[h]
\centering
\caption{\label{tab:classify}
Using off-task classification, we evaluate how standard triplet networks and CSNs affect the convolutional features of the ImageNet-pretrained network they are based on. Naively training a standard triplet network with triplets from different similarity notions hurts the underlying convolutional features.
}
\begin{tabular}{@{}ll@{}} \toprule
    \text{Method} & \text{Top 1 Accuracy}\\ \midrule
    \text{ResNet trained on ImageNet}  & 54.00\% \\ 
    \text{Standard Triplet Network}  & 49.08\% \\ 
    \text{Conditional Similarity Network}  & \text{53.67\%} \\ \bottomrule
\end{tabular}
\end{table}
The results are shown in Table~\ref{tab:classify}. The residual network trained on ImageNet leads to very good initial visual features for general classification tasks. Starting from the pretrained model, we observe that the standard triplet learning approach \emph{decreases} the quality of the visual features, while CSNs retain most of the information. In the triplet prediction experiment in Section~\ref{quant_triplet_pred} standard triplet networks do not perform well, as they are naturally limited by the fact that contradicting notions cannot be satisfied in one single space. This classification result documents that the problem reaches even deeper. The contradicting gradients do not stop at the embedding layer, instead, they expose the entire network to inconsistent learning signals and hurt the underlying convolutional features.

\section{Conclusion}
In this work, we propose Conditional Similarity Networks to learn nonlinear embeddings which incorporate multiple aspect of similarity within a shared embedding. The learned embeddings are disentangled such that each embedding dimension encodes semantic features for a specific aspect of similarity. This allows to compare objects according to various notions by selecting an appropriate subspace using an element-wise mask. We demonstrate that CSNs clearly outperform single triplet networks, and even sets of specialist triplet networks where a lot more parameters are available and each network is trained towards one similarity notion. 

Further, instead of being a black-box predictor, CSNs are qualitatively highly interpretable as evidenced by our exhibition of the semantic submanifolds they learn. Moreover, they provide a feature-exploration mechanism through the learned masks which surfaces the structure of the private and shared features between the different similarity aspects.

Lastly, we empirically find that naively training a triplet network with triplets generated through different similarity notions does not only limit the ability to correctly embed triplets, it also hurts the underlying convolutional features and thus generalization performance. The proposed CSNs are a simple to implement and easy to train end-to-end alternative to resolve these problems.

For future work, it would be interesting to consider learning from unlabeled triplets with a clustering mechanism to discover similarity substructures in an unsupervised way.

\section*{Acknowledgements}
We would like to thank Gunnar R\"atsch and Baoguang Shi for insightful feedback. 
This work was supported in part by the AOL Connected Experiences Laboratory, a Google Focused Research Award, AWS Cloud Credits for Research and a Facebook equipment donation.

{\small
\bibliographystyle{ieee}
\bibliography{egbib}
}

\end{document}